\documentclass{ieeeaccess}
\usepackage{cite}
\usepackage{amsmath,amssymb,amsfonts}
\usepackage{algorithmic}
\usepackage{graphicx}
\usepackage{textcomp}

\usepackage{graphics} 
\usepackage{epsfig} 
\usepackage{mathptmx} 
\usepackage{times} 
\usepackage{amsmath} 
\usepackage{amssymb}  

\usepackage{multirow}
\usepackage{multicol}
\usepackage{tabularx}
\usepackage{threeparttable}
\usepackage[english]{babel}
\usepackage{color}
\usepackage{amsmath}
\usepackage{balance}
\usepackage{caption}
\usepackage{kotex}
\usepackage{url}
\usepackage{hyperref}

\def\BibTeX{{\rm B\kern-.05em{\sc i\kern-.025em b}\kern-.08em
    T\kern-.1667em\lower.7ex\hbox{E}\kern-.125emX}}
\begin{document}
\history{Date of publication xxxx 00, 0000, date of current version xxxx 00, 0000.}
\doi{}

\title{Experimental Evaluation of Precise Placement of the Hollow Object with Asymmetric Pivot Manipulation}
\author{\uppercase{Jinseong Park}\authorrefmark{1}, \IEEEmembership{Member, IEEE},
\uppercase{Jeong-Jung Kim}\authorrefmark{1}, \IEEEmembership{Member, IEEE},\\
\uppercase{and Doo-Yeol Koh}\authorrefmark{1}, \IEEEmembership{Member, IEEE}}
\address[1]{ Department of Artificial Intelligence Machinery, Korea Institute of Machinery \& Materials, Daejeon 34103, of Korea}
\tfootnote{
This research was supported by the National Research Council of Science \& Technology as part of the project entitled ``Developed core technologies for the Robotic General Purpose Task Artificial Intelligence (RoGeTA) framework to enable a variety of everyday services'' (NK254G) and the Industrial Strategic Technology Development Program (no. 20018745 and no. 00416440) funded by the Ministry of Trade, Industry \& Energy (MOTIE, Korea).}

\markboth
{J. Park \headeretal: Experimental Evaluation of Precise Placement of the Hollow Object with Asymmetric Pivot Manipulation}
{J. Park \headeretal: Experimental Evaluation of Precise Placement of the Hollow Object with Asymmetric Pivot Manipulation}

\corresp{Corresponding author: Jinseong Park (e-mail: jspark2090@kimm.re.kr).}

\begin{abstract}
In this paper, we present asymmetric pivot manipulation for picking up rigid hollow objects to achieve a hole grasp. The pivot motion, executed by a position-controlled robotic arm, enables the gripper to effectively grasp hollow objects placed horizontally such that one gripper finger is positioned inside the object's hole, while the other contacts its outer surface along the length. 
Hole grasp is widely employed by humans to manipulate hollow objects, facilitating precise placement and enabling efficient subsequent operations, such as tightly packing objects into trays or accurately inserting them into narrow machine slots in manufacturing processes.
Asymmetric pivoting for hole grasping is applicable to hollow objects of various sizes and hole shapes, including bottles, cups, and ducts. We investigate the variable parameters that satisfy the force balance conditions for successful grasping configurations. Our method can be implemented using a commercially available parallel-jaw gripper installed directly on a robot arm without modification. Experimental verification confirmed that hole grasp can be achieved using our proposed asymmetric pivot manipulation for various hollow objects, demonstrating a high success rate. Two use cases, namely aligning and feeding hollow cylindrical objects, were experimentally demonstrated on the testbed to clearly showcase the advantages of the hole grasp approach.
\end{abstract}


\titlepgskip=-15pt

\maketitle

\section{Introduction}
\label{sec:introduction}
\begin{figure*}[t]
	\centering
	\includegraphics[clip,width=1.00\textwidth]{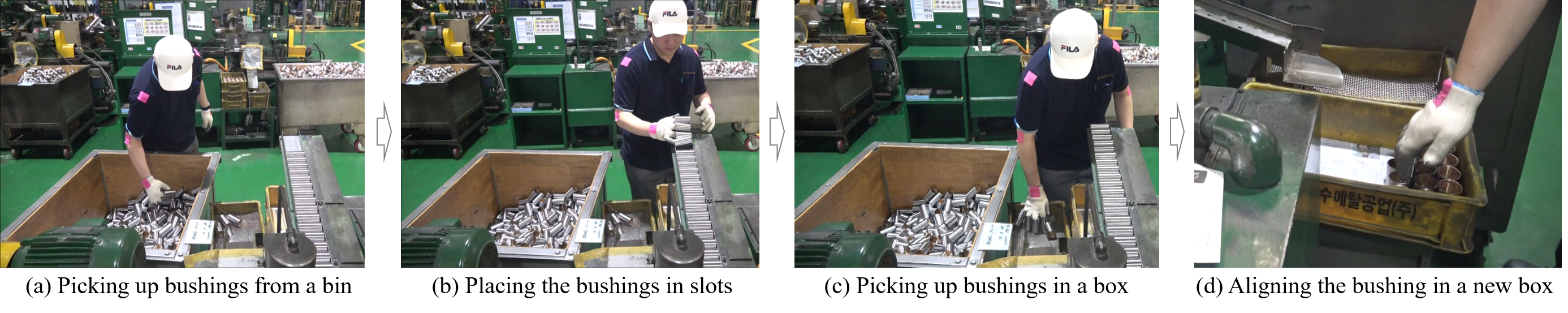}
\caption{Entire manipulation process in the target scenario.}
	\label{fig1}
\end{figure*}

\PARstart{S}{uccessful} grasping of objects in desired poses is essential for precise robotic operations. This paper is concerned with the robotic grasping of objects that has various sizes and types of hole shapes, such as circle hole shapes (bushing, bottle, cup, or can), rectangular hole shapes (duct or mounting rail). The use of hole of the object can be more efficient for various tasks. Indeed, humans commonly use hole grasp for hollow object manipulation to get advantages in subsequent tasks such as transferring many objects in one hand, packing to tray with tight space or feeding to narrow machine slot as depicted in Fig. \ref{fig1}. 

The target configuration of hole grasp is that one finger is positioned inside the object's hole and the other contacts the outer surface, mimicking the way humans grasp hollow objects.
The motivation for adopting a hole grasp approach arises from the need to enhance the feasibility of robotic automation in real-world manufacturing processes. In our previous research, we implemented the target scenario using a pushing primitive based on Cartesian force control \cite{Park}. 
To ensure precise placement into narrow machine slots, we utilized the slot's flat surface as an external pusher, but feedback from machining process manager indicated that consistent impacts could compromise machine reliability and durability.
In response to these concerns, we have embraced a hole grasp methodology that more closely mimics human motion.
The most straightforward approach for achieving hole grasp is to lay the gripper on the ground to align with the object's hole using a large range of arm motions and then grasp the hole directly; however, this method may require excessive free space for arm movements as illustrated in Fig. \ref{fig13}(a). Sequential procedure with pinch grasping the lying object, placing it in the vertical direction and grasping it again with the hole can be another approach \cite{Tournassoud}. 
However, these conventional methods require regrasping, which increases tact time and is therefore undesirable for manufacturing automation \cite{Monkman}. The use of suction gripper is also not suitable for hole grasp since sufficient suction area cannot be guaranteed in the presence of the hole.

On the other hand, in-hand manipulation with dexterity can be exploited. This method can be categorized based on whether it utilizes resources intrinsic to the hand (intrinsic dexterity) or extrinsic to the hand (extrinsic dexterity). From the perspective of intrinsic dexterity, in-hand manipulation with finely position- and force-controlled articulated hands has been developed to achieve the dexterity and flexibility of human hands \cite{Salisbury, Wimbock, Yoshikawa}.
Conversely, extrinsic dexterity with a simple gripper leverages external factors such as gravity, contact forces, and dynamic arm motions to perform push, sliding, and pivot primitives \cite{Dafle1}. 
For practical applications, employing a simplified end effector can provide a more effective and appropriate solution than a gripper with complex mechanisms.

Pivot manipulation has been actively explored through various approaches and perspectives. In essence, a passive end effector can be utilized to achieve pivot manipulation by leveraging gravity \cite{Mucchiani}.
An open-loop robotic arm trajectory was found to rotate an object using a pinch grasp without considering form/force closure consideration \cite{Holladay}. 
Pivoting by prehensile pushing against the environment based on arm motion planning and a frictional contact model has been formulated \cite{Dafle2}. A feedback control based pivoting grasp has also been studied. The pivoting of the object was achieved by using gravity and adaptive control of the gripping forces with vision/tactile \cite{Vina} or force/tactile \cite{Costanzo} feedback. Swing-up regrasping has been implemented by means of dynamic arm motions based on the energy control approach \cite{Sintov}. 
The non-prehensile pivoting utilizing contacts of environment was implemented with hybrid force-velocity control \cite{Hou}. 
Sequential pivoting operation was presented to move an heavy object on the floor as if it walking \cite{Aiyma}.
Learning based pivoting grasp also has been implemented. A deep predictive model with tactile sensing was designed for the active slip control of an end effector in the presence of gravity or acceleration \cite{Stepputtis}. Q-learning based open loop pivoting was performed using inertial forces generated from the robot arm motion \cite{Cruciani}. 
Implementation of pivot manipulation from a mechanical customization perspective has also been conducted.
A mechanically improved parallel-jaw gripper has been developed to gain advantages in pivot-like grasping. A gripper consisting of one thumb and one finger based on epicyclic gear train was designed to achieve scooping without robot arm motion \cite{Babin}. A gripper equipped with sheet embedded belt enabled passively pull-in operation during closing the gripper to flip the object \cite{Morino}. A fingertip with pneumatic braking mechanism realized pivoting a grasped object with aligning in gravity direction \cite{Taylor}. 
However, most of pivoting manipulations have been realized by symmetric pinch grasping or a parallel-jaw gripper customization and, targeted in rigid and solid objects.

The applications for thin objects grasping is more relevant to our purpose from the perspective of the asymmetric grasping configuration and the use of environment contact for secure grasping. A scooping grasp for picking up thin objects lying on a flat surface has been proposed in which one (another) finger makes contact with the bottom (top) of a thin object while executing a scooping-like sliding motion \cite{Levesque, Levesque}. Tilting thin objects lying on flat surface with respect to edge in contact with environment was realized by customized fingernail for secure grasping \cite{Tong}. An in-hand manipulation for inserting thin, peg-like parts into shallow holes has been proposed based on tilt and regrasping primitives \cite{Kim}. 
However, to the best of our knowledge, the use of an object's hole represents the first attempt in pivot manipulation.

This paper investigates asymmetric pivot manipulation of hollow objects using a simple gripper, focusing on their unique geometric characteristics.
Experiments were performed with a two-fingered parallel-jaw gripper installed on a robot arm without additional modification. Hollow cylinders and prisms of various sizes and shapes were tested to verify the effectiveness of our technique. Two industrial applications were designed to highlight the advantages of our asymmetric pivoting method. One is packing process that align objects by placing repetitively in vertical direction to tray with tight space. For tight packing, the force sensor installed on wrist was used to adjust the contact force with the tray and already packed objects. Second is feeding task that insert objects into machine slots having a narrow clearance by positioning the gripper on the tip of the slot and dragging the object off the gripper.

The main contributions of this work are twofold, which can be summarized as follows.
\begin{enumerate}
\item	The desired grasp configuration of the hollow object, termed hole grasp, is achieved through asymmetric pivot manipulation, which is facilitated by position control of the robotic arm and a conventional parallel-jaw gripper.

\item	The automation of machining processes, traditionally performed by humans manually, is achieved using the proposed approach, enabling consistent feeding of objects into narrow machine slots and aligning them with package boxes within tight spaces, and has been validated on a real testbed.
\end{enumerate}

The rest of this paper is organized as follows. In Sec. II, our problem description is presented. The proposed asymmetric pivot for hole grasp is presented in Sec. III, and the experimental evaluation to real testbed is presented in Sec. IV. Finally, Sec. V presents the conclusions.

\section{Problem Description}

\subsection{Target Scenario}
The target manufacturing process shown in Fig. \ref{fig1} involves machining hollow cylindrical objects, such as bushings
The detailed steps of the process are to (a) pick up objects from a bin by inserting one of the fingers of the hand into the hole in the object, (b) drop the object in a machine slot, (c) pick up the machined object with the identical grasping configuration as in (a), and (d) align it on a tray in a longitudinal direction and then repeat the process. It seems that the target process is very simple for a human worker but, it is rather challenging from the perspective of robotic manipulation. While inserting a finger into the target object's hole offers advantages in scenarios involving multiple objects, with one grasped on each finger, this paper focuses solely on single-object grasping due to the use of a 1-DOF, two-finger gripper.

\subsection{Problem Statement}
The hollow object grasping addressed in this paper is quasi-static, as it does not rely on the object's dynamic motion or inertial forces. While pivot manipulation is often achieved through gravity or dynamic arm motion, our approach ensures that it is performed at a sufficiently slow speed.
Hole grasping can be considered as a two-dimensional plane motion because once gripper is closed to complete initial configuration, one finger of gripper caught in the hole is mostly suppress off the plane motion.
The bin environment is approximated as a flat surface, and the object lies in an arbitrary orientation.
An object of interest is rigid objects that can have various sizes and shapes of hole. 
Because the finger of the gripper in contact with outer surface of the object necessarily push the object to make secure grasping by maintaining contact between vertex of the object and the ground, asymmetric pivoting of flexible hollow objects, such as bottles or rubber tubes, may have a low success rate due to their tendency to crumple; the limitation is experimentally demonstrated in sec. IV.

\begin{figure}[t]
	\centering
	\includegraphics[clip,width=1.0\columnwidth]{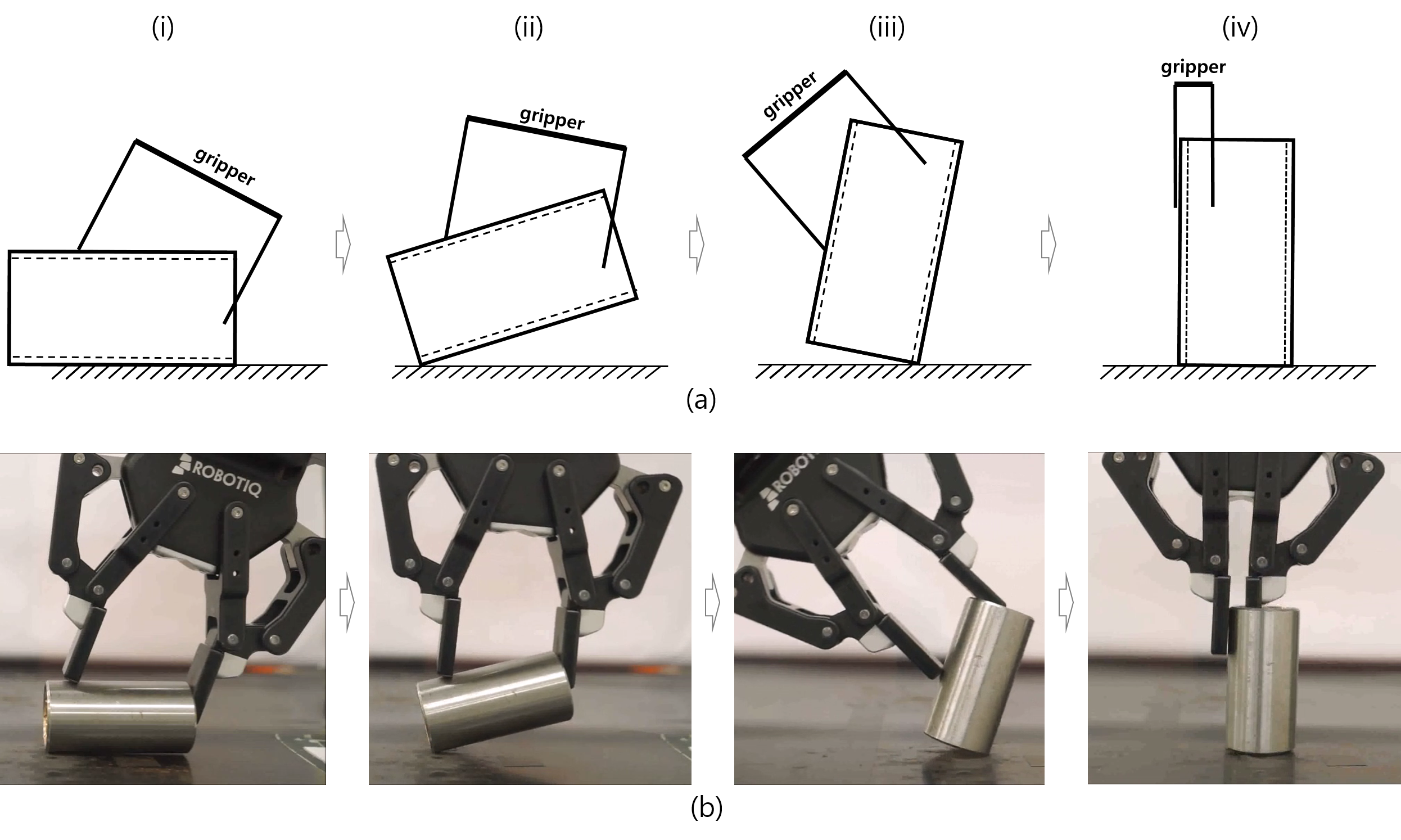}
    \caption{In-hand manipulation of a hollow object with a desired configuration: (a) schematic and (b) real implementation.}
	\label{fig2}
\end{figure}

Based on the defined problem statement, the target scenario can be formulated as the asymmetric pivot manipulation problem.
The whole procedure of asymmetric pivot manipulation is depicted in Fig. \ref{fig2}. First, a hollow object is initially lying in the horizontal direction and the gripper approaches to the object with a small tilt angle. By closing the gripper, one finger is in contact with the upper surface of the object, and the other is in contact with the tip of the inner surface of the hole (i). The center of rotation is the vertex of the opposite side that is in contact with ground (ii). The robot arm pivots the object from the horizontal to the vertical direction with respect to the center of rotation (iii). Finally, the desired grasp configuration is achieved by inserting one of the gripper while the other gripper maintains the contact with outer surface (iv). Therefore, the  hole  grasp  can  be achieved  by  completing  a  single  manipulation  process. Also, a quasi-static force balance can be achieved, although the configuration is asymmetric, as one of the gripper outside the hole pushes the object during pivoting to maintain the contact between object and surface. 

\section{Asymmetric Pivot Manipulation for Hole Grasp}
The feasible configurations are investigated based on three parameters describing the planar model and the friction coefficients at each contact. 
On this basis, the entire grasping maneuver for a hollow object is performed in three steps.

\begin{figure}[t]
	\centering
	\includegraphics[clip,width=0.9\columnwidth]{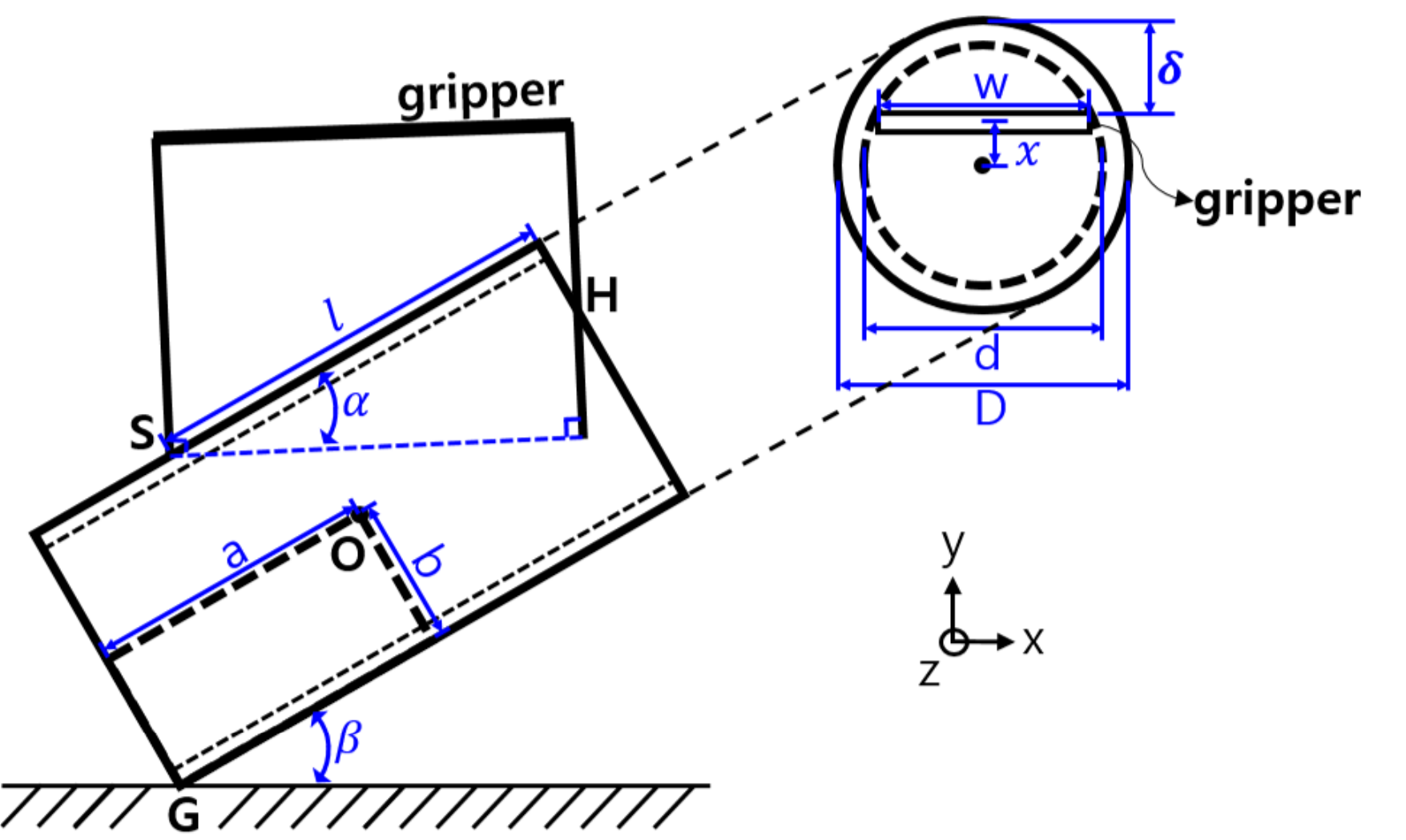}
\caption{Parameters and configuration for the manipulation of a hollow object.}
	\label{fig3}
\end{figure}

\begin{table}[t]
    \centering
    \caption{Parameters considered for the manipulation of a hollow object.}
    \label{table1}
    \begin{threeparttable}
    \begin{tabular}{c l}
    \hline
        Parameter & Description \\
    \hline
        \textit{l}\tnote{\dag} & Distance between S and the corner of the object\\
        $\alpha$\tnote{\dag} & Angle between the object and the gripper \\
        $\beta$\tnote{\dag} & Angle between the object and the ground \\
        $\delta$ & Distance between H and the corner of the object \\
        a & Half length of the object \\
        b & Half height of the object \\
        D & Outer diameter of the object \\
        d & Inner diameter (hole diameter) of the object \\
        w & Width of the gripper \\
        x & Distance between the gripper and the center of the hole  \\
    \hline
    \end{tabular}
    \begin{tablenotes}[para]
        \item[\dag] Variable.
    \end{tablenotes}
    \end{threeparttable}
\end{table}
\setlength{\textfloatsep}{3mm}  

\begin{figure}[t]
	\centering
	\includegraphics[clip,width=0.75\columnwidth]{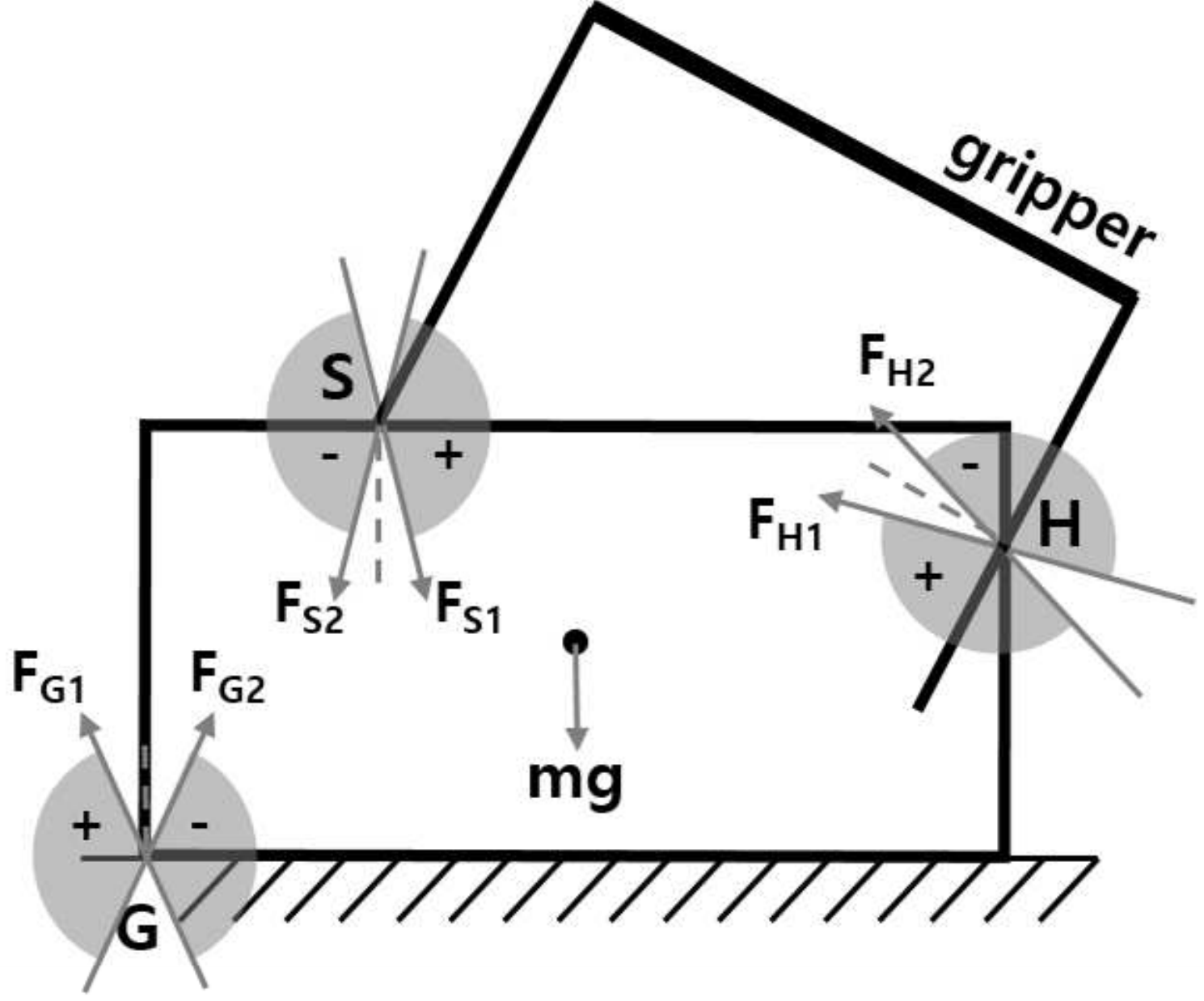}
\caption{Schematic diagram of the unit contact wrenches and moment labeling for contacts S, H and G.}
	\label{fig4}
\end{figure}

\begin{table}[t]
    \centering
    \caption{Basis contact wrenches for contact points S, H and G.}
    \label{table2}
    \begin{tabular}{c| c | l }
    \hline
        \multicolumn{2}{c|}{Items} & Values \\
    \hline
        \multirow{6}{*}{S} & \multirow{3}{*}{S1} & $m_{S1} = (\textit{l}-a)\cos{\gamma_S} - b \sin{\gamma_s}$ \\ 
        & & $F_{S1,x} = \sin{(\beta+\gamma_S)}$ \\ 
        & & $F_{S1,y} = -\cos{(\beta+\gamma_S)}$ \\ \cline{2-3}
        & \multirow{3}{*}{S2} &  $m_{S2} = (\textit{l}-a)\cos{\gamma_S} + b \sin{\gamma_s}$ \\ 
        & & $F_{S2,x} = \sin{(\beta-\gamma_S)}$ \\ 
        & & $F_{S2,y} = -\cos{(\beta-\gamma_S)}$ \\ 
    \hline
        \multirow{6}{*}{H} & \multirow{3}{*}{H1} & $m_{H1} = a\sin{(\alpha - \gamma_H)} + (b-\delta) \cos{(\alpha - \gamma_H)}$ \\ 
        & & $F_{H1,x} = -\cos{(\alpha - \beta - \gamma_H)}$ \\ 
        & & $F_{H1,y} = \sin{(\alpha - \beta - \gamma_H)}$ \\ \cline{2-3}
        & \multirow{3}{*}{H2} & $m_{H2} = a\sin{(\alpha + \gamma_H)} + (b-\delta) \cos{(\alpha + \gamma_H)}$ \\ 
        & & $F_{H2,x} = -\cos{(\alpha - \beta + \gamma_H)}$ \\ 
        & & $F_{H2,y} = \sin{(\alpha - \beta + \gamma_H)}$ \\ 
    \hline
        \multirow{6}{*}{G} & \multirow{3}{*}{G1} & $m_{G1} = -a\cos{(\gamma_G - \beta)} - b\sin{(\gamma_G - \beta)}$ \\ 
        & & $F_{G1,x} = -\sin{\gamma_G}$ \\
        & & $F_{G1,y} = \cos{\gamma_G}$ \\ \cline{2-3}
        & \multirow{3}{*}{G2} & $m_{G2} = -a\cos{(\gamma_G + \beta)} + b\sin{(\gamma_G + \beta)}$\\ 
        & & $F_{G2,x} = \sin{\gamma_G}$ \\ 
        & & $F_{G2,y} = \cos{\gamma_G}$ \\ 
    \hline
    \end{tabular}
\end{table}
\setlength{\textfloatsep}{3mm}  

\subsection{Variable Parameters}
The two-fingered parallel-jaw gripper and the ground surface form three contact points with the target object, denoted by S, H and G, in the presence of Coulomb friction, as illustrated in Fig. \ref{fig3}. 
From these contact points, three variable parameters representing the grasping configuration defined from the object's perspective can be extracted, as shown in Table \ref{table1}: \textit{l}, $\alpha$ and $\beta$. The remaining seven parameters in Table \ref{table1} are all constants.
The contact point S, i.e., the position at which one finger is in contact with the upper surface of the object, is determined by adjusting the length of the gripper stroke as it approaches the object. To obtain a more general mathematical description, a nondimensional parameter $l_a = l/(2a)$ is introduced, which can take values in the following range:
\begin{eqnarray}
	\label{eq_1}
    \begin{array}{l}
    0 < l_a \leq 1
    \end{array}
\end{eqnarray}
The position of the contact point H is related to a constant parameter x that is determined by the relationship between the gripper width, w, and the hole diameter, d. This parameter x is represented as follows:

\begin{eqnarray}
	\label{eq_2}
    \begin{array}{l}
    x = \frac{d}{2}\sin{(\cos^{-1}{(\frac{w}{d})})}=\frac{d}{2}\sqrt{1-(\frac{w}{d})^2}
    \end{array}
\end{eqnarray}
where $0 < \frac{w}{d} < 1$ because the gripper width cannot exceed the hole diameter if finger insertion is to be successful. Consequently, $\delta$ can be determined from Eq. \eqref{eq_2} as follows:

\begin{eqnarray}
	\label{eq_3}
    \begin{array}{l}
    \delta = \frac{D}{2} - x
    \end{array}
\end{eqnarray}

\begin{figure*}[t]
	\centering
	\includegraphics[clip,width=1.00\textwidth]{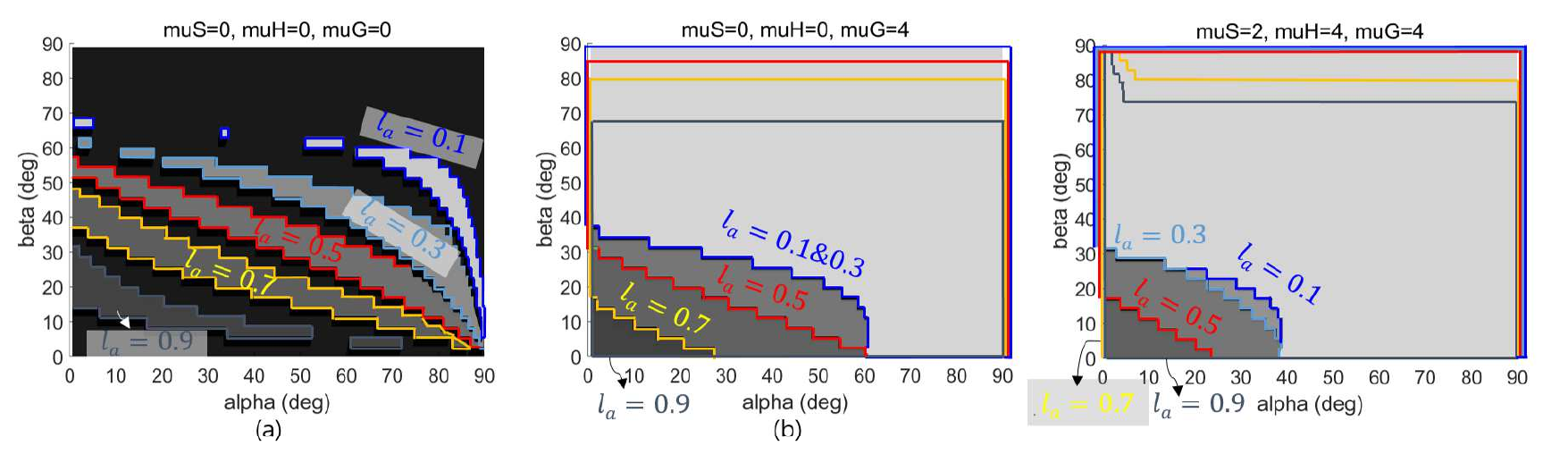}
\caption{Stable grasp configurations for a bushing (see Fig. \ref{fig9}) are represented in the $\alpha\beta$-plane as areas with varying $l_a$ values for three cases of the friction coefficients: (a) all contacts are frictionless; (b) only the ground contact has friction, with $\mu_G$=0.4; and (c) all three contact points have friction, with $\mu_S$=0.2, $\mu_H$=0.4 and $\mu_G$=0.4. Five lines in different colors represent increasing values of $l_a$. In all cases, an increase in friction expands the stable grasp region in both $\alpha$ and $\beta$. Meanwhile, an increase in $l_a$ extends the stable grasp region toward the origin ($\alpha=\beta=0$) while reducing its coverage of higher values of $\beta$.}
	\label{fig5}
\end{figure*}

\begin{figure}[t]
	\centering
	\includegraphics[clip,width=1.0\columnwidth]{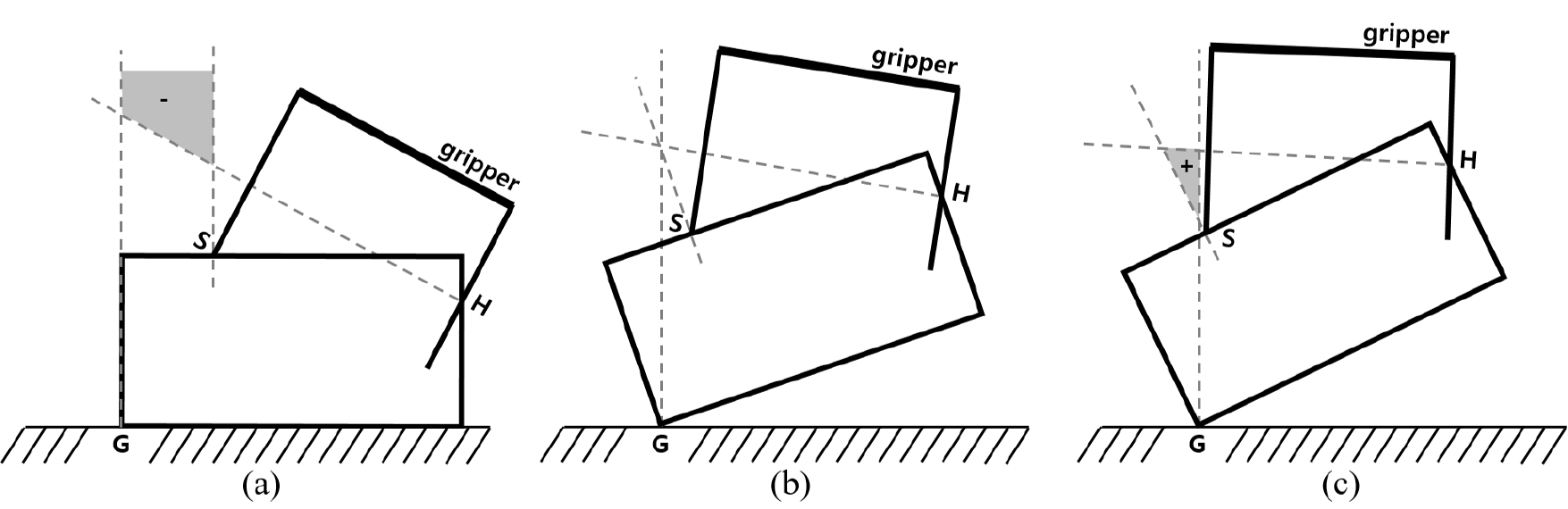}
\caption{Kinematic contact analysis ($\mu_G=\mu_s=\mu_H=0$). Among the possible combinations of the variable parameters, only a few grasp configurations yield first-order form closure: (a) In the initial position, $\beta=0$, we can never achieve form closure. b) A form-closure grasp is available within a small range of $\beta$. (c) Form closure becomes impossible again with slightly more rotation.}
	\label{fig6}
\end{figure}

The variable angles $\alpha$ and $\beta$ lie in the following ranges:
\begin{eqnarray}
	\label{eq_4}
    \begin{array}{l}
    0 < \alpha <  \frac{\pi}{2}
    \end{array}
\end{eqnarray}
\begin{eqnarray}
	\label{eq_5}
    \begin{array}{l}
    0 \leq \beta \leq  \frac{\pi}{2}
    \end{array}
\end{eqnarray}

where there is no equality in Eq. \eqref{eq_4}. In the case of $\alpha = 0$, the grasp is no longer an asymmetric pivot configuration for hole grasp, but pinch grasp. Because the gripper cannot be caught in the hole, the contact point H will be positioned at the upper right corner of the object in Fig. \ref{fig3}. In addition, Eq. \eqref{eq_3} no longer holds; rather, $\delta$ becomes zero. In the extreme opposite case, $\alpha = \pi/2$, means that the gripper is aligned with the hole of the object by considerable arm motion and the hole is grasped directly. This situation is exactly what we want to avoid because it requires a great deal of free space and may even be impossible to achieve when the object is placed in a small bin, the sides of which will cause interference. Note that in this case, the distance \textit{l} is no longer an independent variable; rather, it becomes a constant of $\textit{l} = \delta$ due to the corresponding grasping configuration. Our goal is to grasp the object with as small an $\alpha$ as possible to minimize interference with the external environment. As $\beta$ denotes the tilt angle of the object with respect to the ground, its value ranges from 0, indicating that the object is lying horizontally on the ground, to $\pi/2$, indicating a complete pivot to the vertical direction, as represented in Eq. \eqref{eq_5}.

\subsection{Stable Grasp Modeling}
The unit contact wrenches for each contact point S, H and G and the gravitational force are illustrated in Fig. \ref{fig4}.
The friction coefficients at the contact points are $\mu_s$, $\mu_H$, and $\mu_G$, respectively, and the friction angles, which are the half angles of the corresponding friction cones, are calculated as $\gamma_S=\tan^{-1}{\mu_s}$, $\gamma_H=\tan^{-1}{\mu_H}$ and $\gamma_G=\tan^{-1}{\mu_G}$. In the presence of friction, we have the chance to achieve force-closure grasping; for example, the grasp configuration illustrated in Fig. \ref{fig4} is in force closure since the total set of wrenches, namely, the composite wrench cone, does not share any wrench space, and therefore, the external wrench of the gravitational force can be balanced \cite{Lynch}. Even if the grasp is not in force closure, we still have the chance to achieve a force balance against external wrenches with the desired grasp configuration.

The values of the three variable parameters and the contact friction coefficients necessary to achieve the desired force balance are investigated under quasi-static conditions by mathematically formulating all of the unit contact wrenches as represented in Table \ref{table2}. Each contact point has two contact wrenches, each of which has three elements in the planar model, as follows:
\begin{eqnarray}
	\label{eq_6}
    \begin{array}{l}
    F_{i} = 
    \begin{bmatrix}
    m_i \\
    F_x \\
    F_y
    \end{bmatrix}
    \end{array}
\end{eqnarray}
where $i = S1, \cdot\cdot\cdot, G2$. 
The force balance configurations are analyzed by solving the corresponding linear programming problem in simulation and are presented in Fig. \ref{fig5}, based on the combinations of the variables $l$, $\alpha$ and $\beta$ for three friction coefficient sets, including the frictionless condition.
When all contacts are frictionless, the problem is defined as a first-order form-closure problem, as follows:
\begin{eqnarray}
	\label{eq_7}
    \begin{array}{l}
    min \quad \sum{k_i} \\
    s.t \ \ \quad  \sum{k_i F_i} = 0 \\
        \ \ \quad \quad k_i \geq 1
    \end{array}
\end{eqnarray}

\begin{figure}[t]
	\centering
	\includegraphics[clip,width=1.0\columnwidth]{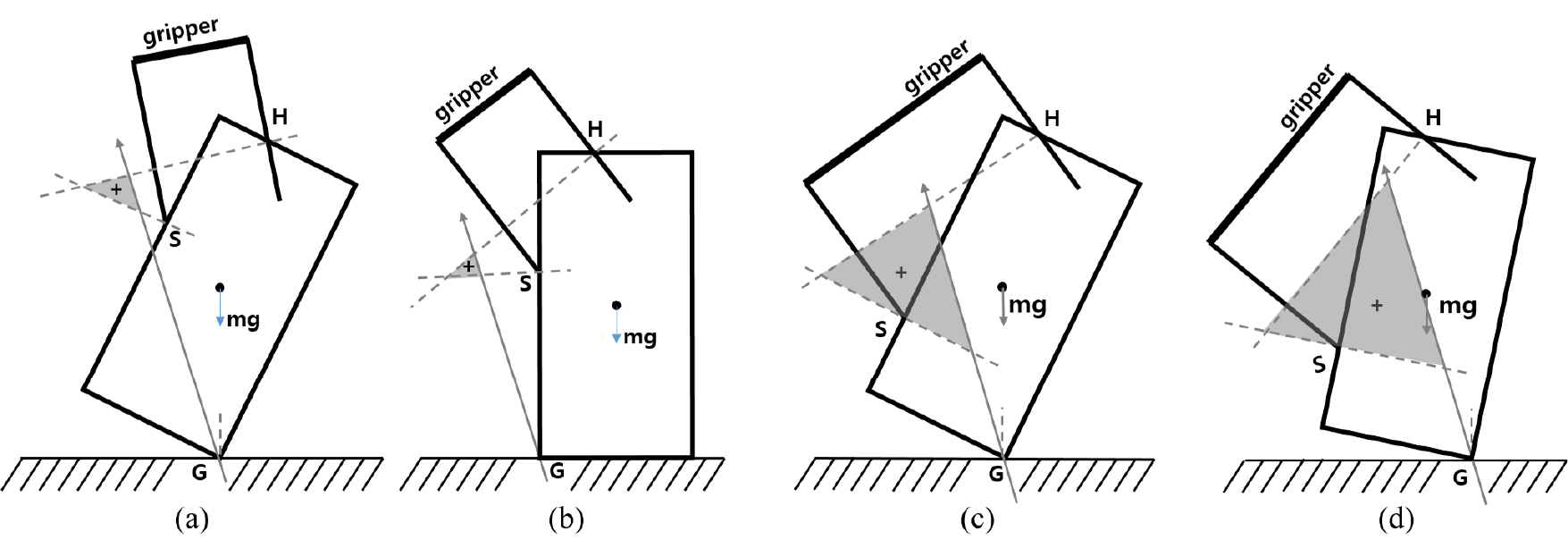}
\caption{Force balance analysis during the manipulation task with $\mu_G=0.4$ and $\mu_S=\mu_H=0$. (a, b) In the case of $l_a \leq 0.5$, the force balance is maintained throughout the whole range of $\beta$. (c, d) In the case of $l_a > 0.5$, the external wrench from the gravitational force cannot be balanced when $\beta > \beta_{ub}$.}
	\label{fig7}
\end{figure}

Without friction, a stable grasp for pivot manipulation is difficult to attain. In the initial position $\beta=0$, using our grasp configuration, undesirable in-hand motion of the object cannot be fully prevented, as shown in Fig. \ref{fig6}(a); indeed, there is no stable grasp configuration at $\beta=0$ in the $\alpha\beta$-plane. The only way to achieve a form-closure grasp at $\beta=0$ is to apply a pinch grasp, which directly implements a hole grasp with $\alpha=90^{\circ}$. In other positions, a form-closure grasp appears in a few cases (see Fig. \ref{fig6}(b)), but with slightly more rotation, it is lost again (Fig. \ref{fig6}(c)).
In the presence of friction, on the other hand, the following force balance problem is considered:
\begin{eqnarray}
	\label{eq_8}
    \begin{array}{l}
    min \quad \sum{k_i} \\
    s.t \ \ \quad  F_{ext} + \sum{k_i F_i} = 0 \\
        \ \ \quad \quad k_i \geq 0
    \end{array}
\end{eqnarray}

\begin{figure}[t]
	\centering
	\includegraphics[clip,width=0.9\columnwidth]{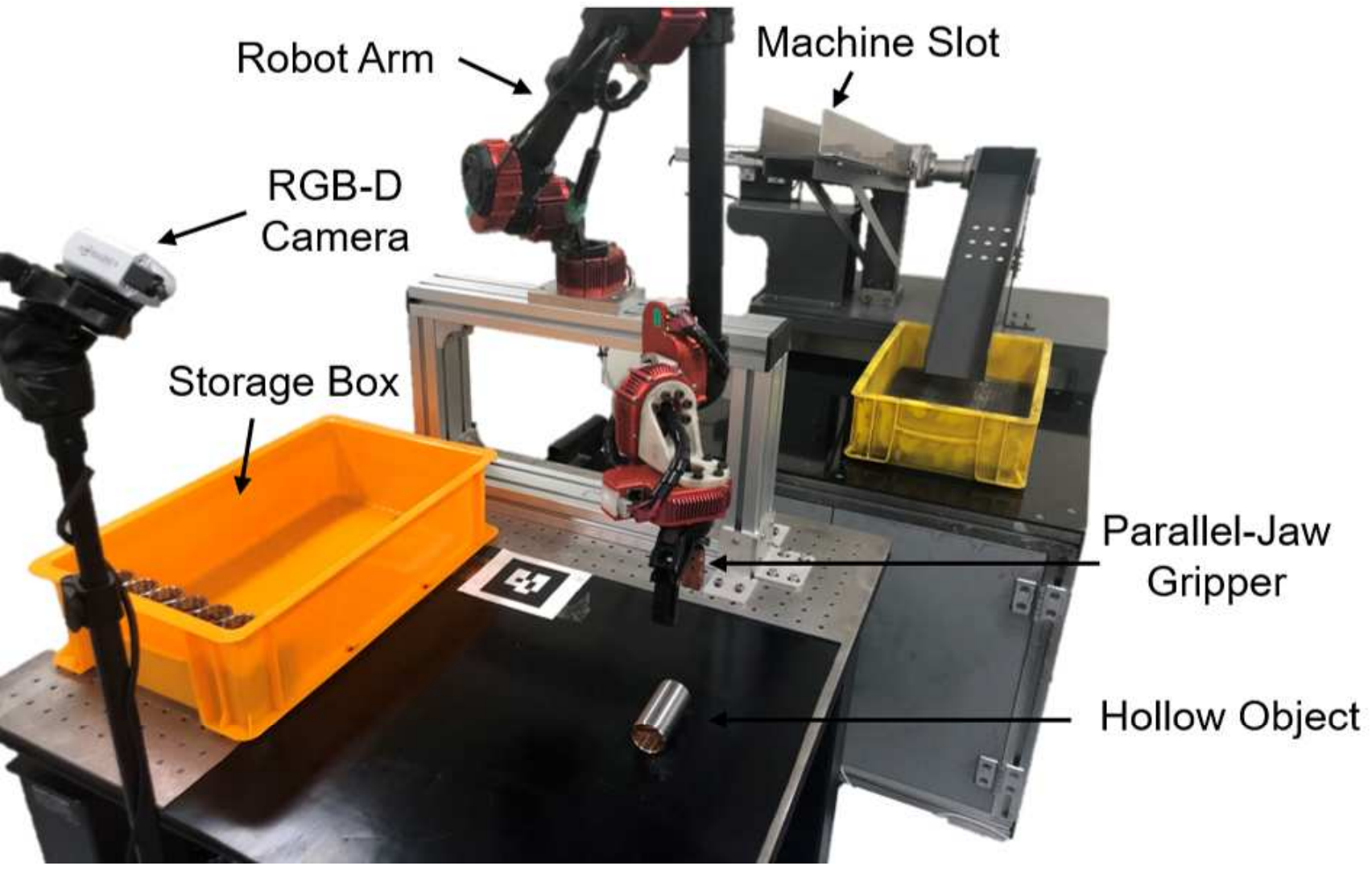}
\caption{Experimental setup with a 1-DOF gripper attached to a 6-DOF robotic arm, an RGB-D camera, a hollow object and the work space.}
	\label{fig8}
\end{figure}

\noindent where $F_{ext}=[0, 0, -mg]^T$. The dominant parameters that can be adjusted to expand the stable grasp region are the ground contact friction $\mu_G$ and the distance $l_a$, as seen by comparing Fig. \ref{fig5}(a) and (b).
An increase in $\mu_G$ expands the stable grasp region in both $\alpha$ and $\beta$; the desired configuration now yields a force balance in the initial position $\beta=0$.
An increase in $l_a$ enables the establishment of a stable grasp configuration with a smaller $\alpha$.

However, as $l_a$ increases, the stable grasp region eventually becomes limited by an upper bound on $\beta$, namely, $\beta_{ub}$. Fig. \ref{fig7} illustrates the two cases: with a small $l_a$, the line of the gravitational force does not cross the consistent positive moment throughout the whole $\beta$ range (Fig. \ref{fig7}(a) and (b)), whereas with a sufficiently large $l_a$, the gravitational force vector intersects with the consistent positive moment when $\beta > \beta_{ub}$ (Fig. \ref{fig7}(c) and (d)). $\beta_{ub}$ is a function of the geometrical parameters and the contact friction and can be obtained numerically from Eq. \eqref{eq_8}.
In the presence of contact friction with the ground, a finite $\beta_{ub}$ is present only when the center of mass of the object crosses the ground contact point G ($\beta_{ub} > \tan^{-1}{(a/b)}$).
Therefore, only a positive moment is generated, which is helpful for pivoting the object to the vertical direction.
Furthermore, the object will not escape from the gripper even if the force balance is broken because one finger is already inserted into the hole.
The presence of friction at points S and H also helps to expand the stable grasp region, as shown in Fig. \ref{fig5}(c).

The results can be summarized as follows: (a) at least the contact between the object and the ground should have sufficient stiction to ensure a stable grasp, especially in the initial position, and (b) from a simulation of the stable grasp configurations, the feasible relationship between $\alpha$ and $l_a$ that permits manipulation can be determined.

\subsection{Grasping Maneuver for a Hollow Object}
The proposed maneuver for grasping a hollow object with a two-fingered gripper consists of three steps, as illustrated in Fig. \ref{fig1}.

\begin{itemize}

\item \textbf{Grasp} the object with a tilt angle $\alpha$ such that one finger is caught in the hole in the object to make contact at point H while the other makes contact at S (Fig. \ref{fig2}(i)).
\item \textbf{Tilt} the object with respect to point G while maintaining the necessary force balance (Fig. \ref{fig2}(ii) and (iii)).
\item \textbf{Align} the gripper longitudinally parallel to the object to complete the grasping maneuver (Fig. \ref{fig2}(iv)).

\end{itemize}

Our goal is to grasp the object with as large an $l_a$ as possible to permit the use of a small $\alpha$ (see Fig. \ref{fig5}).

\section{Implementation}
The proposed hole grasp method was experimentally verified for hollow cylinders of various sizes and shapes as well as hollow prisms. Two application examples were also investigated to demonstrate the usefulness of our method.

\begin{figure}[t]
	\centering
	\includegraphics[clip,width=0.8\columnwidth]{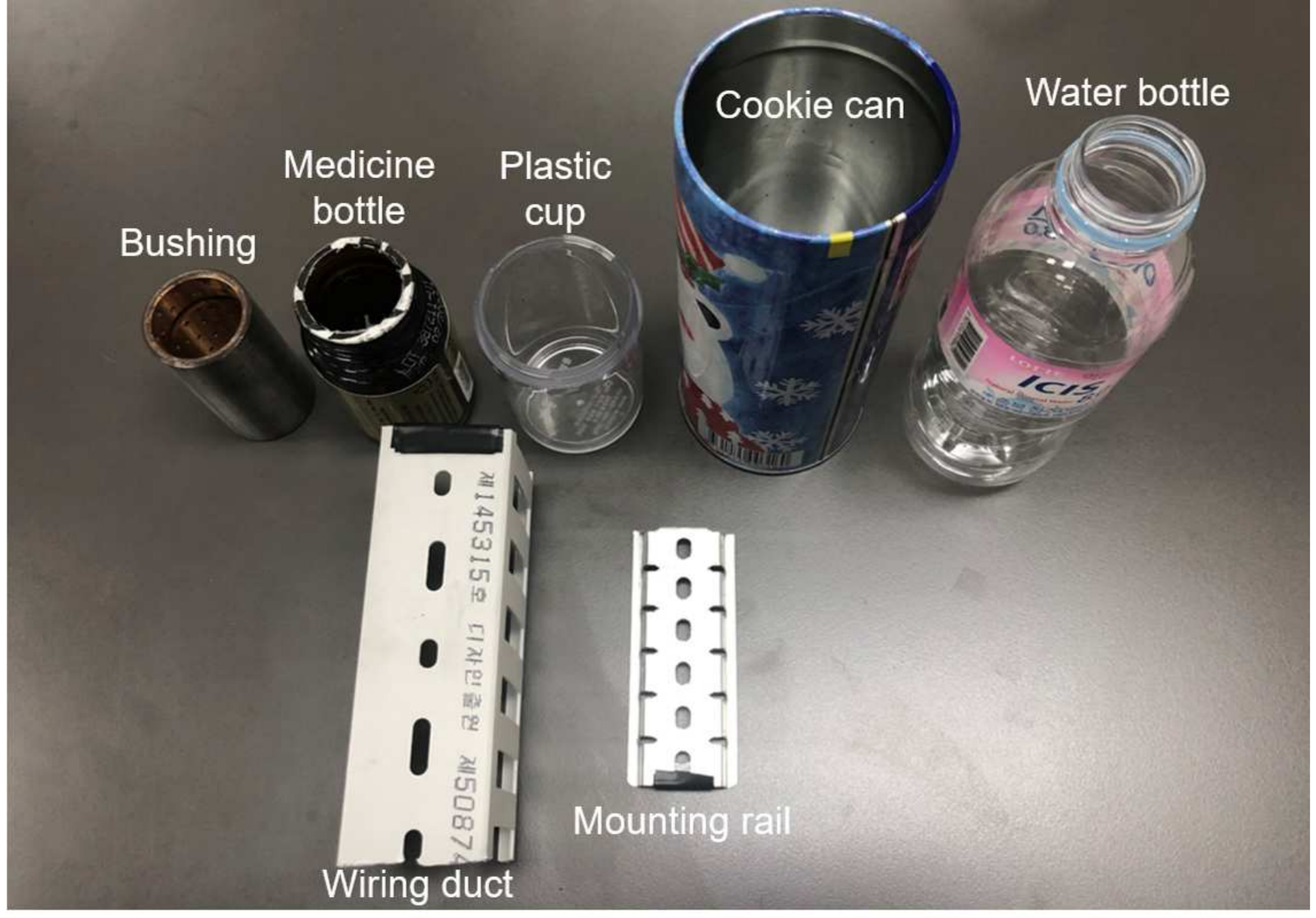}
\caption{Hollow objects of various sizes and shapes.}
	\label{fig9}
\end{figure}

\subsection{Experimental setting}
The hardware setup used in the experiments is shown in Fig. \ref{fig8}.
A simple 1-DOF wire-driven gripper with 2 rigid fingers and common finger tips was attached to a 6-DOF serial robotic arm from HEBI Robotics that consists of 6 individual series elastic actuators (SEAs) \cite{Rollinson}. The hardware components communicated by means of the Robot Operating System (ROS).
An externally equipped RGB-D camera (Intel RealSense D435) is used to detect the 3D pose of the object, with the Mask R-CNN algorithm \cite{He}, \cite{Matterport} modified to incorporate depth information.With an April tag attached to the surface, the relative pose between the origin of the robot arm and the object can be calculated.

\begin{table*}[ht]
    \centering
    \caption{Experimental results for the manipulation of hollow objects.}
    \label{table3}
    \begin{threeparttable}
    \begin{tabular}{p{2.5cm} p{2.5cm} p{2cm} p{1cm} p{1cm} p{1.5cm} p{1.5cm} p{1.5cm}}
    \hline\hline
        \multirow{2}{*}{Hollow object} & Dimensions & $l_a$ & $\alpha$ & $\delta$ & Success rate  & 95\% CI & 95\% CI \\
        & (l$\times$D$\times$d)~[mm] & & [rad] & [mm] & [$\%$]  & (Lower) & (Upper)\\
    \hline
        Bushing         & 68  $\times$ 34  $\times$ 28  & 0.9$\rightarrow$0.65 & $\pi$/10  & 7.2  & 10/10 & 72.25\% & 100.0\% \\[0.5ex]
        Medicine bottle     & 83  $\times$ 35  $\times$ 28  & 0.75$\rightarrow$0.63   & $\pi$/10  & 13.2 &  9/10 & 59.58\% & 98.21\%  \\[0.5ex]
        Plastic cup  & 73  $\times$ 48  $\times$ 44  & 0.85$\rightarrow$0.38 & $\pi$/10  & 5.9  &  8/10  & 49.02\% & 94.33\%  \\[0.5ex]
        Cookie can    & 160 $\times$ 66  $\times$ 60  & 0.28                  & $\pi$/3.8 & 4.7  & 10/10  & 72.25\% & 100.0\%  \\[0.5ex]
        Wiring duct     & 122 $\times$ 60  $\times$ 57  & 0.53                  & $\pi$/10  & 3    & 10/10   & 72.25\% & 100.0\%  \\[0.5ex]
        Mounting rail   & 83  $\times$ 7.6 $\times$ 6.4 & 0.67                   & $\pi$/20  & 1.2  &  3/10  & 10.74\% & 60.27\%  \\[0.5ex]
        Water bottle    & 161 $\times$ 30  $\times$ 25  & 0.28                  & $\pi$/4   & 20.5 &  0/10 & 0.0\% & 26.46\%  \\
    \hline\hline
    \end{tabular}
    \end{threeparttable}
\end{table*}
\setlength{\textfloatsep}{3mm}  

To realize application, mock-up hardware is also constructed to represent the bushing machining process. A machine slot with a 5 mm slot clearance, which allows for placement only when the bushing is well positioned, was placed in front of the robotic manipulator. After machining, the machined bushing needed to be aligned in a box. The the whole scenario of experiments is available at \url{https://www.youtube.com/watch?v=4fnZTY9C2SI}.

\begin{figure}[t]
	\centering
	\includegraphics[clip,width=1.0\columnwidth]{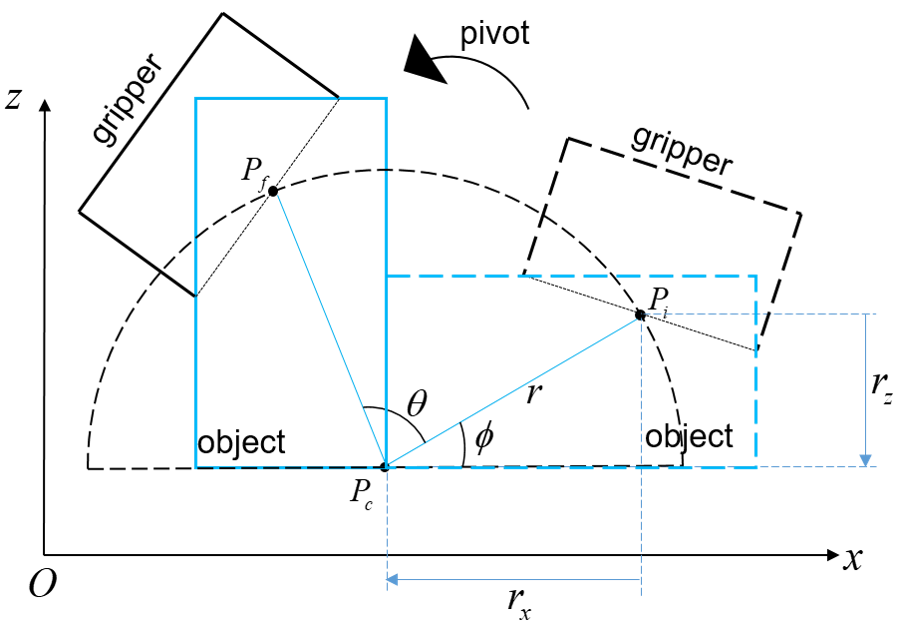}
    \caption{Trajectory generation for asymmetric pivot manipulation}
	\label{fig10}
\end{figure}

\subsection{Experimental Results}
The main target objects considered in the experiments were rigid hollow cylinders of various sizes and types, including a bushing, a medicine bottle (glass), a plastic cup, and a cookie can (steel); a hollow prism, such as a wiring duct, was also tested, as shown in Fig. \ref{fig9}.
A nonrigid object (a water bottle) and a protruding thin object (a mounting rail) were also tested to clarify the limits of our technique. 

Experimental results along with the corresponding object information and grasp configuration parameters are represented in Table \ref{table3}.
The trajectory for asymmetric pivot manipulation is generated as illustrated in Fig. \ref{fig10}.
Once the target objective is assigned, the first step involves determining the gripper's tilt angle $\alpha$ and distance $l_a$ based on the results shown in Fig. \ref{fig5}. The gripper's center is then positioned at $P_i$ to achieve the initial grasp, ensuring that one end of the gripper securely grasps the object's hole while maintaining a gripping distance greater than $l_a$. A pivot motion is executed around $P_c$ with radius $r$ and an angle $\theta$. To orient the object vertically, $\theta$ is typically set to 90$^{\circ}$. If sliding occurs, as noted by changes in $l_a$ in Table \ref{table3}, $\theta$ is adjusted to less than 90$^{\circ}$.

\begin{figure*}[t]
	\centering
	\includegraphics[clip,width=1.35\columnwidth]{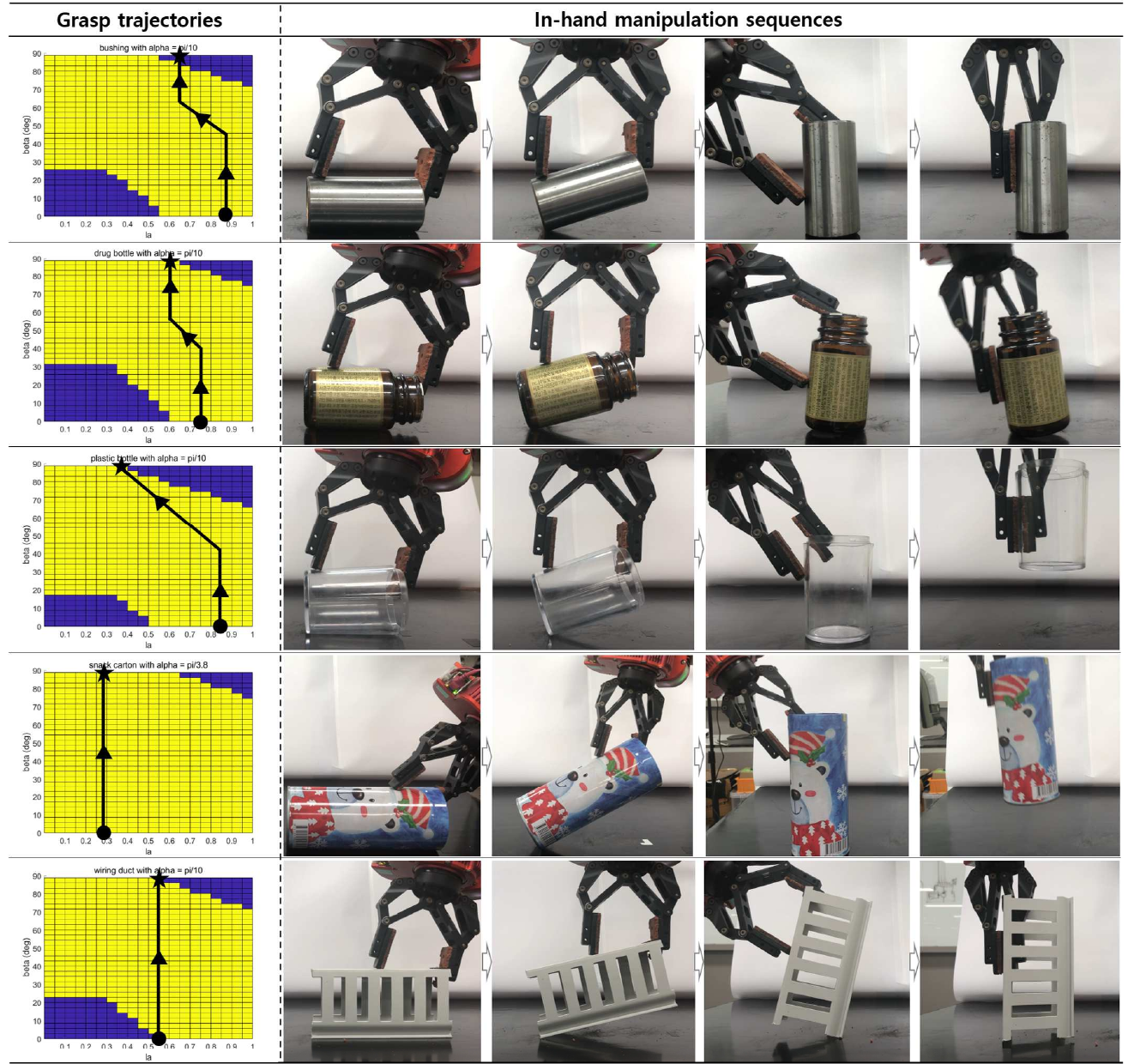}
\caption{Experimental results with grasp trajectories and in-hand manipulation sequences for successful cases: (a) a bushing, (b) a medicine bottle, (c) a plastic cup, (d) a cookie can and (e) a wiring duct. In each grasp trajectory plot, the friction coefficient set used is $\mu_S$=0.2, $\mu_H$=0.4 and $\mu_G$=0.4.} The yellow region represents the stable grasp region, and the circle and star marks indicate the initial and final grasp configurations, respectively.
	\label{fig11}
\end{figure*}

\begin{figure*}[t]
	\centering
	\includegraphics[clip,width=1.35\columnwidth]{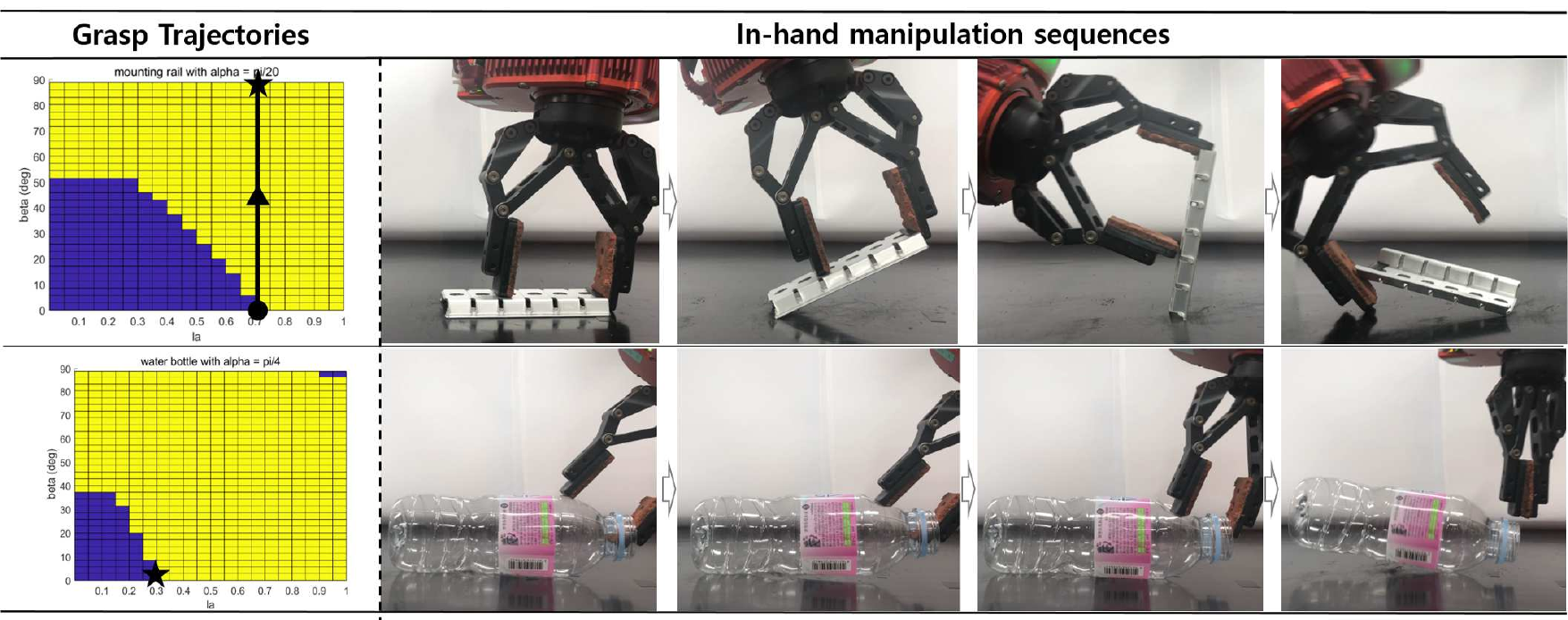}
\caption{Experimental results with grasp trajectories and in-hand manipulation sequences for unsuccessful cases: (a) a mounting rail and (b) a water bottle. In each grasp trajectory plot, the friction coefficient set used is $\mu_S$=0.2, $\mu_H$=0.4 and $\mu_G$=0.4. The yellow region represents the stable grasp region, and the circle and star marks indicate the initial and final grasp configurations, respectively.}
	\label{fig12}
\end{figure*}

For each object, 10 trials were performed; the overall success rate was 71.4$\%$, and manipulation was successful only for the target hollow objects, excluding the nonrigid bottle and the protruding thin object, with a corresponding success rate of 94$\%$.
Fig. \ref{fig11} (Fig. \ref{fig12}) shows the asymmetric pivot manipulation process for several successful (unsuccessful) cases. Note that the grasp trajectory plots, where the x- and y-axes represent $l_a$ and $\beta$, respectively, illustrate how the grasp configuration changes throughout the manipulation process.
For the grasp trajectories, the friction coefficient set used is $\mu_S$=0.2, $\mu_H$=0.4 and $\mu_G$=0.4, which was inferred from experimental results through multiple iterative experimental adjustments as a representative set that best explains the experimental outcomes.

The bushing was grasped with a perfect success rate. During the \textbf{Tilt} maneuver, the contact at point S naturally transitioned into a sliding contact in all trials..
Sliding at S causes $l_a$ to decrease, as seen from the grasp trajectory plot in Fig. \ref{fig11}. Consequently, the force balance is maintained over a long range of $\beta$ and is eventually broken near the end of the rotation; due to the interference from the finger in the hole, the desired pose is achieved even if the contact at G breaks.
The medicine bottle and plastic cup each have a small entrance and thus a large $\delta$ parameter. For these objects, failure occurred in a total of three trials as a result of out-of-plane motion due to the slippery surface of the object. The grasp trajectory also changed during the \textbf{Tilt} maneuver because of the sliding contact at point S.
A long hollow cylinder could also be pivoted using our in-hand manipulation method. Because our gripper had an 82 mm stroke, the $l_a$ for the cookie can could not be long, meaning that a relatively large $\alpha$ was necessary to achieve a stable grasp. In this case, no sliding occurred during manipulation.
The wiring duct was also successfully manipulated with the same method.

The thin mounting rail was grasped only 3 times because the object could not maintain its vertical posture after the successful completion of the \textbf{Tilt} maneuver.
The water bottle was never successfully grasped in the desired configuration due to the crushing of the surface around contact point S, resulting in undesirable sliding at contact point H. Note that, alternatively, to achieve hole grasping for non-rigid or protruding thin objects, modifications the gripper’s fingertip with a fingernail-like attachment and employing scooping manipulation may improve the success rate.

To validate the consistency and reliability of our experimental results, we conducted a Wilson Score Confidence Interval (95\%) analysis, as it provides a more accurate confidence interval estimation for small sample sizes compared to the normal approximation method.
To calculate the 95\% confidence interval (CI) for each success rate, we applied the following Wilson Score formula:
\begin{eqnarray}
	\label{eq_9}
    \begin{array}{l}
    \hat{p}_{\text{adjusted}} = \frac{\hat{p} + \frac{Z^2}{2n} \pm Z \sqrt{\frac{\hat{p}(1 - \hat{p})}{n} + \frac{Z^2}{4n^2}}}{1 + \frac{Z^2}{n}}
    \end{array}
\end{eqnarray}
where $\hat{p}$ is observed success rate, $n$ is total number of trails, $Z$ is critical value from the standard normal distribution (1.96 for 95\% level), respectively. The results are represented in Table \ref{table3}.

The results of the experiments, each conducted with 10 trials, show that objects with a 100\% success rate, such as the Bushing, Cookie can, and Wiring duct, have a lower bound of the confidence interval above 72\%, indicating a strong likelihood of consistent performance. For objects with a success rate above 80\%, such as the Medicine bottle and Plastic cup, the confidence intervals exhibit some degree of variability, yet still indicate a high probability of success, confirming that the technique remains reliable within the tested conditions. Consequently, despite some variation in the success rates among the five rigid objects tested, the results from the 10-trial experiments indicate a relatively reliable level of success.

On the other hand, objects with low success rates, such as the Mounting rail and Water bottle, exhibit significantly wider confidence intervals, indicating greater uncertainty in the outcomes. In particular, for the non-rigid object, where the success rate is extremely low, reducing this uncertainty would require a significantly larger number of trials.
Nevertheless, beyond statistical analysis, the inclusion of the non-rigid object was intended to demonstrate the inherent difficulty of achieving a hole grasp using the proposed method. The low success rate and the corresponding wide confidence interval quantitatively reinforce our conclusion, further highlighting the challenge of applying our technique to non-rigid objects.

\begin{figure}[t]
	\centering
	\includegraphics[clip,width=1.0\columnwidth]{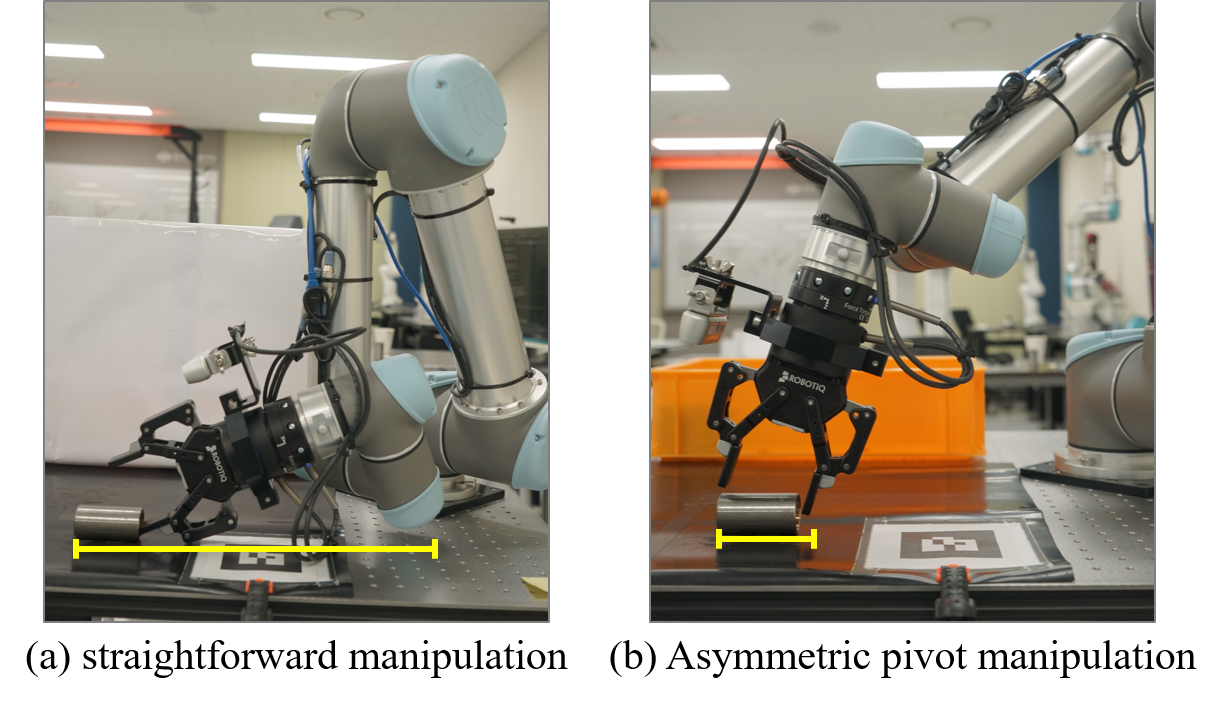}
    \caption{Comparison of required footprint sizes for hole grasp of the hollow object: (a) straightforward manipulation (b) proposed asymmetric pivot manipulation}
	\label{fig13}
\end{figure}

\subsection{Applications}
Two application examples were designed based on bushing machining processes that are commonly performed by humans in industry, as shown in Fig. \ref{fig1}, to demonstrate the usefulness of our target grasping configuration.
For the application realization, an alternative robotic arm (Universal Robots UR5) equipped with a 6-axis FT sensor (Robotiq FT300s) and a two-finger gripper (Robotiq 2F-85) is used instead of SEA based robots to better address practical commercial scenarios, as shown in Fig. \ref{fig13}. 
The FT sensor was used to monitor the reaction force between the object and the ground during pivot manipulation, initiating an emergency stop when the force exceeded a predefined threshold.
Admittance \cite{Keemink} or impedance \cite{Park2024} controllers could be considered for ensuring safe interaction with the ground during pivot manipulation. However, to enable straightforward application in real-world scenarios, the implementation was carried out using position control in Cartesian space, a feature available in all commercial robotic arms.

\begin{figure*}[t]
	\centering
	\includegraphics[clip,width=2.00\columnwidth]{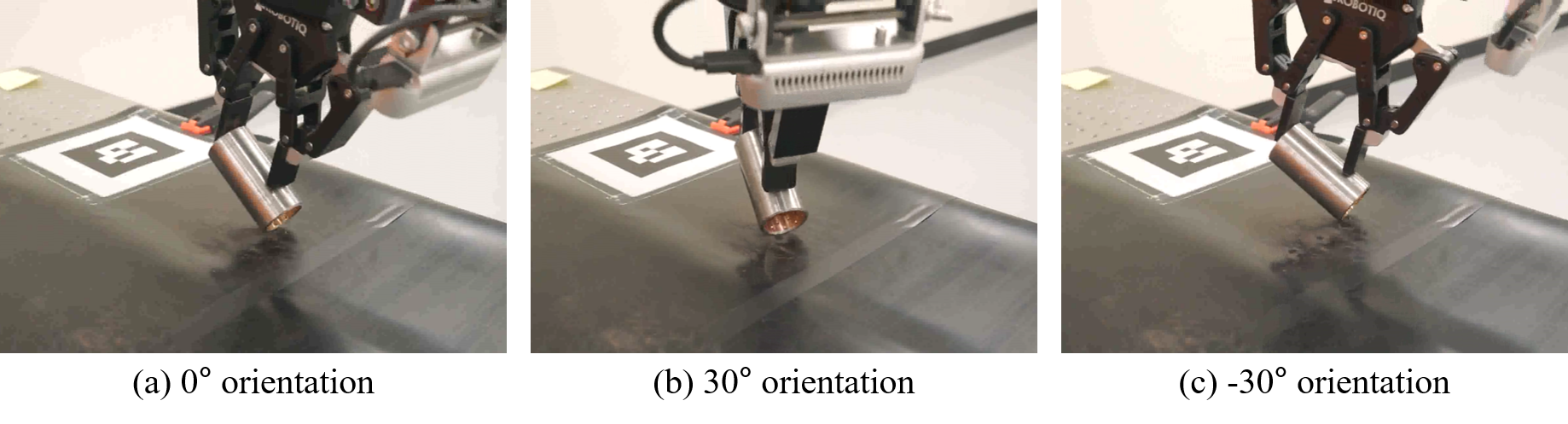}
\caption{pivot manipulation with various orientation}
	\label{fig14}
\end{figure*}
\begin{figure*}[h]
	\centering
	\includegraphics[clip,width=2.00\columnwidth]{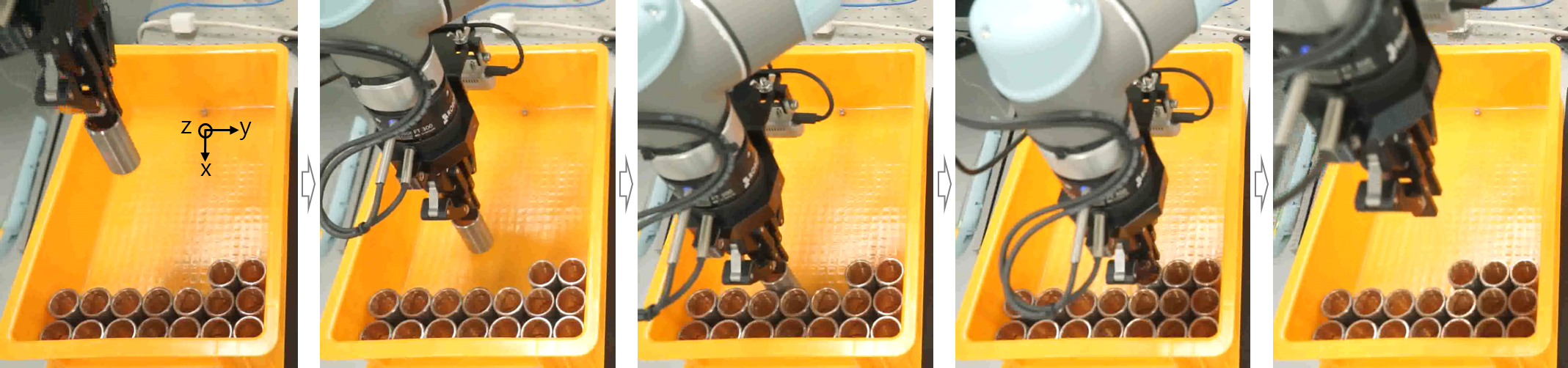}
\caption{Sequences for bushing alignment in a storage box.}
	\label{fig15}
\end{figure*}
\begin{figure*}[h]
	\centering
	\includegraphics[clip,width=2.00\columnwidth]{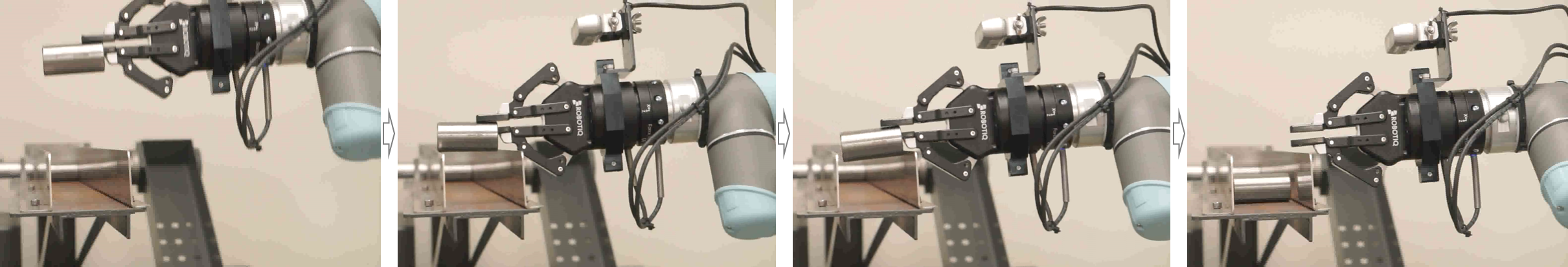}
\caption{Sequences for bushing insertion into a machine slot.}
	\label{fig16}
\end{figure*}

The proposed asymmetric pivot manipulation for hole grasp requires a much smaller footprint than the typical pinch grasp, making it more effective for practical worksites or bins as shown in Fig. \ref{fig13}. 
Additionally, the RGB-D sensor, rather than being equipped externally, was mounted on the end-effector. Tus, the horizontal orientation of the target object is estimated using the Mask R-CNN algorithm, and the pivot manipulation is performed accordingly, as shown in Fig. \ref{fig14}.

The first example is aligning a bushing in the vertical direction in a storage box. A hybrid force and position control for guiding the interaction of the robot arm with the external environment was designed to achieve tight alignment in order to maximize the storage capacity \cite{Park}. Fig. \ref{fig15} shows sequences corresponding to bushing alignment: the bushing is pushed sequentially in the negative direction along the z-axis, in the positive direction along the x-axis and in the positive direction along the y-axis until it reaches the surface of the box or the previously aligned bushings. Thanks to our hole grasp configuration, the bushing can be pushed securely into target position; furthermore, no regrasping is required during the alignment process.

The second example is inserting an object into a machine slot with a tight clearance, as shown in Fig. \ref{fig16}. Specifically, our scenario consists of placing a bushing by (a) placing the gripper finger that is inside the hole in the object into the slot, (b) releasing the gripper force and (c) dragging the gripper away so that the bushing is released from the finger, analogous to the operation being performed by the human worker in Fig. \ref{fig1}(b). Although a symmetric pinch grasp can be used as an alternative grasp for realizing insertion into a slot with a tight clearance, a pinch grasp requires much more accurate grasping in the face of a number of difficulties, such as imprecise sensing, uncertainties in the planning and control of the robot arm and interference from neighboring objects. By contrast, our method does not require accurate grasping; it requires only the successful establishment of the target grasp configuration.

\subsection{Discussion}
While $\mu_G$ maintains the stability of the grasp, $\mu_H$ and $\mu_S$ always have the potential to slide. Among the six basis contact wrenches, $F_{H1}$ and $F_{S2}$ rarely contribute to achieving force balance; the values of $k_{H1}$ and $k_{S2}$ in the solution to Eq. \eqref{eq_8} are very small or zero over the entire simulation range. Indeed, in the experiments with $l_a>0.5$ (bushing, medicine bottle and plastic cup), sliding at S always occurred.
Out-of-plane motion was the most significant reason for failure, except in the cases of the protruding thin object and the nonrigid object. Undesirable z-axis motion can occur for a cylindrical object when the contact at S is not secure because the contact area is less than that for a prismatic object. Modifying the fingertip of the gripper to be stickier would help prevent motion along the z-axis. Our method requires the target object to be a rigid body because the grasp configuration depends on the security of the contact with the object's surface. The object should also be able to stand on its own in the vertical position between the \textbf{Tilt} and \textbf{Align} maneuvers; therefore, a thin object is not suitable for manipulation using our technique.

Our target grasp configuration shows the following advantages in the considered application examples: (1) it enables secure pushing of an object during interaction with the external environment, and (2) it relieves the need for accurate grasping for tasks such as placing an object into a slot compared with symmetric pinch grasping, for which the grasping position needs to be relatively accurate.
Although the motion of the robot arm is important for maintaining contact with the ground and the gripper during the manipulation process, a high-performance robot arm is not required for our approach; the accuracy of the robot arm used in our experiments, Sec. IV.B, is reported to be $\pm$2 cm by the manufacturer.

\section{Conclusions}
A robotic in-hand manipulation method is essential for attaining the desired grasp configuration for an object that is not initially placed in the desired pose. We present asymmetric pivot manipulation for a hollow object held by a two-fingered parallel-jaw gripper that allows a horizontally placed object to be grasped such that one finger is positioned inside the hole in the object and the other is in contact with the outer surface in a longitudinal manner, thus offering an advantage for subsequent operations, such as placing or aligning the object in a desired configuration. Our manipulation technique can be applied to hollow objects of various sizes and shapes by analyzing the stable grasp configurations based on simulations with variable parameters. Experiments with various target objects yield a high success rate, and the limitations of our approach, such as the requirement of object rigidity, are also clarified. Two application examples, namely, aligning an object in a box and placing an object in a machine slot, demonstrate the contributions of the proposed in-hand manipulation method.
The proposed asymmetric pivot manipulation-based hole grasping technique holds significant potential for advancements in robotic automation and manufacturing processes, as precise placement tasks are often more critical than accurate picking. Our study differentiates itself by focusing on precise grasping for object placement, whereas most existing research primarily emphasizes picking alone.

In our future work, it will be necessary to investigate manipulation in dynamic environments, including the handling of randomly stacked objects and uneven surfaces. In practice, the objects of interest will be randomly piled in a bin, causing the target object to be positioned at a tilted angle. Additionally, the object will not be stably in contact with the ground but will instead interact irregularly with its surroundings or other objects.
Furthermore, planar simplification may no longer be appropriate, necessitating analyses of stable grasping and grasp maneuvers in spatial cases. To facilitate the practical implementation of the proposed method, feedback control, achieved through motion detection with a tactile sensor on the fingertip and an AI-based detection algorithm, will be essential in future work.

\begin{IEEEbiography}[{\includegraphics[width=1in, height=1.25in, clip,keepaspectratio]{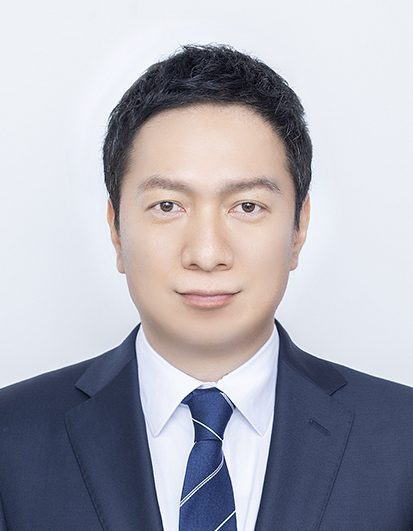}}]
    {Jinseong Park} (Member, IEEE) received B.S., M.S. and Ph.D. degrees in mechanical engineering from the Korea Advanced Institute of Science and Technology (KAIST), Daejeon, Republic of Korea, in 2008, 2010 and 2016, respectively. He was a senior researcher at Hyundai Robotics Co., Ltd., Ulsan, Republic of Korea, from 2016-2017.
    He has been a senior researcher at the Korea Institute of Machinery and Materials (KIMM), Daejeon, Republic of Korea, since 2017. His research interests include optimal control, active vibration control, and fault diagnosis, with an emphasis on mobile manipulators, vehicles, and magnetic levitation systems.
\end{IEEEbiography}

\begin{IEEEbiography}[{\includegraphics[width=1in, height=1.25in, clip,keepaspectratio]{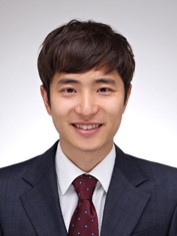}}]
    {Jeong-Jung Kim} (Member, IEEE) received B.S. in electronics and information engineering from
Chonbuk National University, Jeonju, Korea, in 2006 and the M.S. degree in robotics and Ph.D. degree in electrical engineering from the Korea Advanced Institute of Science and Technology (KAIST), Daejeon, Korea,  2008, 2015, respectively. He was a postdoctoral researcher at Korea Institute of Science and Technology (KIST), Seoul, Republic of Korea, from 2015 to 2017. He has been a senior researcher at the Korea Institute of Machinery and Materials (KIMM), Daejeon, Republic of Korea, since 2017, and a principal researcher there since 2024. His research interests include AI for manipulation and navigation in robotics.

\end{IEEEbiography}

\begin{IEEEbiography}[{\includegraphics[width=1in, height=1.25in, clip,keepaspectratio]{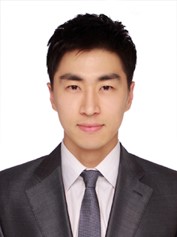}}]
    {Doo-Yeol Koh} (Member, IEEE) received B.S. in Ajou University, Suwon, Republic of Korea in 2006, Ph.D. degrees in mechanical engineering from the Korea Advanced Institute of Science and Technology (KAIST), Daejeon, Republic of Korea, in 2013. He was a principal researcher at Samsung Heavy Industries Co., Ltd., Daejeon, Republic of Korea, from 2013-2017. He has been a senior researcher at the Korea Institute of Machinery and Materials (KIMM), Daejeon, Republic of Korea, since 2017. His research interests include design and control of a modular mobile manipulator, bimanual teleoperation for teaching, and AI-based motion generation.

\end{IEEEbiography}

\EOD

\end{document}